\documentclass[acmsmall]{acmart}
\usepackage{booktabs} % For formal tables
\usepackage{subfigure}
\usepackage{blindtext}
\usepackage{graphicx}
\usepackage{csquotes}
\usepackage{caption}
\usepackage{amsmath}
\usepackage{siunitx}
\usepackage{microtype}
\usepackage{multicol}
\usepackage{upgreek}
\usepackage{booktabs,tabularx,ragged2e,url}
\usepackage[english]{babel}
\usepackage{fancyhdr}
\usepackage[noend]{algpseudocode}
\usepackage{float}
\usepackage{blindtext}

\usepackage{academicons}
\usepackage{xcolor}
\usepackage{nomencl}
\makenomenclature

\usepackage[ruled,linesnumbered,commentsnumbered]{algorithm2e}

\SetAlFnt{\small}
\SetAlCapFnt{\small}
\SetAlCapNameFnt{\small}
\SetAlCapHSkip{0pt}
\IncMargin{-\parindent}

\usepackage[hang,flushmargin]{footmisc}

\usepackage{xcolor,colortbl}
\definecolor{Gray}{gray}{0.90}
\newcolumntype{a}{>{\columncolor{Gray}}c}

\usepackage{xcolor,soul}
\definecolor{light-gray}{gray}{0.95}
\sethlcolor{light-gray}

\graphicspath{{Figures/}}
\newcommand{\fig}[1]{Fig.~\ref{#1}}
\newcommand{\tb}[1]{Table~\ref{#1}}
\newcommand{\eq}[1]{(\ref{#1})}

\usepackage{enumitem,booktabs}
\usepackage[referable]{threeparttablex}
\renewlist{tablenotes}{enumerate}{1}
\makeatletter
\setlist[tablenotes]{label=\tnote{\alph*},ref=\alph*,itemsep=\z@,topsep=\z@skip,partopsep=\z@skip,parsep=\z@,itemindent=\z@,labelsep=.2em,leftmargin=*,align=left,before={\footnotesize}}
\makeatother

\usepackage[font={small}]{caption}

\DeclareCaptionLabelSeparator{periodspace}{.\quad}
\usepackage{enumitem}
\newlist{todolist}{itemize}{2}
\setlist[todolist]{label=$\square$}
\usepackage{pifont}
\AtBeginDocument{%
  }

\begin{document}

\title{Time-Series Forecasting and Sequence Learning Using Memristor-based Reservoir System}

\author{Abdullah M. Zyarah}
\email{abdullah.zyarah@utsa.edu}
\orcid{1234-5678-9012}
\authornotemark[1]
\affiliation{%
  \institution{\\University of Texas at San Antonio, USA, and University of Baghdad}
  \country{Iraq}
}

\author{Dhireesha Kudithipudi}
\affiliation{%
  \institution{\\University of Texas at San Antonio}
  \country{USA}}
\email{dk@utsa.edu}

\begin{abstract}
Pushing the frontiers of time-series information processing in the ever-growing domain of edge devices with stringent resources has been impeded by the systems' ability to process information and learn locally on the device. Local processing and learning of time-series information typically demand intensive computations and massive storage as the process involves retrieving information and tuning hundreds of parameters back in time. In this work, we developed a memristor-based echo state network accelerator that features efficient temporal data processing and in-situ online learning. The proposed design is benchmarked using various datasets involving real-world tasks, such as forecasting the load energy consumption and weather conditions. The experimental results illustrate that the hardware model experiences a marginal degradation in performance as compared to the software counterpart. This is mainly attributed to the limited precision and dynamic range of network parameters when emulated using memristor devices. The proposed system is evaluated for lifespan, robustness, and energy-delay product. It is observed that the system demonstrates reasonable robustness for device failure below 10\%, which may occur due to stuck-at faults. Furthermore, 247$\times$ reduction in energy consumption is achieved when compared to a custom CMOS digital design implemented at the same technology node.
\end{abstract}





\maketitle

\section{Introduction}
Empowering edge devices with the capability to process and learn continuously locally on the device is essential when handling stationary and non-stationary time-series data, as remote cloud services have proven to be slow and unsecured~\cite{kim20192}. While such an objective enhances reliability and adaptability to edge constraints, it is challenging to satisfy in devices with limited resources. In this work, the local processing and learning from time-series data is explored within the context of reservoir networks, particularly echo state network (ESN)~\cite{jaeger2001echo}. ESN is developed to handle continuous data of stationary and non-stationary nature and has powered a large number of applications, including prosthetic finger control~\cite{zyarah_esn_2023}, stock market prediction~\cite{wang2021stock}, cyber-attack and anomaly detection~\cite{hamedani2017reservoir, ullah2022intelligent}, weather forecasting~\cite{de2023forecasting}, and modeling dynamic motions in bio-mimic robots~\cite{soliman2021modelling}. The network features efficient temporal data processing and unrivaled training speed~\cite{cerina2020echobay, bianchi2020reservoir}. Furthermore, it does not suffer from the known training problems in most recurrent neural networks, such as vanishing and exploding gradients, which occur due to the recurrent connections and gradient backward propagation through time~\cite{zhang2023survey}. Thus, running ESN on edge devices using dedicated custom-designed hardware not only enables the algorithm to make temporal decisions in real-time but also yields a significant improvement in power efficiency and throughput~\cite{kudithipudi2016design,gauthier2021next}.

These reasons urge the research community to investigate the physical implementation of ESN. There are several digital~\cite{honda2020hardware, kleyko2020integer, gan2021cost} and mixed-signal implementations of the ESN networks in the literature using memristors devices, MOSFET switching, and photonics~\cite{sorokina2020multidimensional, kume2020tuning, wlaźlak2020neuromorphic, zhang2023survey}. However, this work will place emphasis on the memristor-based implementations, owing to their numerous advantages over other approaches, such as the integration with CMOS devices and fabrication using the standard foundry process~\cite{jiang2019integrating}. Furthermore, the memristor intrinsic non-linearity feature enhances the reservoir separability and echo state properties~\cite{moon2019temporal}. A recent attempt at memristive ESN, including a hardware implementation of the ESN network using a memristor double-crossbar array, was presented by Hassan et al. in 2017.. The proposed design is evaluated for autonomous signal generation using the Mackey-Glass dataset and the simulation is done in MATLAB. In 2019, Wen et al. presented a memristor-based ESN trained in online fashion using least mean square~\cite{wen2018memristor}. The network is evaluated for short-term power load forecasting. In the same year, a memristor-based reservoir computing system, which leverages the concept of the virtual node, was introduced by Moon at al.~\cite{moon2019temporal}. The virtual node concept involves building a single reservoir node and using it to virtually emulate a chain of other nodes in the reservoir. The proposed system features high-power efficiency but suffers from high latency and short lifespan, and it is demonstrated in speech recognition and time-series forecasting. In 2022, the same concept was adopted to build a cyclic reservoir computing system based on memristor devices. The proposed design is implemented in a 65nm technology node and verified on hand-written vowel recognition task~\cite{liang2022rotating}. Lately, Nair et al. proposed a memristive ESN with an extended synaptic sampling machine (SSM) to ensure synaptic stochasticity and power efficiency~\cite{nair2023essm}. The proposed neural system is benchmarked for classification tasks using ECG, MNIST, Fashion MNIST, and CIFAR10.

While a few of the ESN implementations are equipped with in-situ learning to endow the ESN network with fast learning and adaptation capabilities, none extend the learning to cover the structural plasticity (synaptogenesis), which is imperative for controlling the network's sparsity level. To the best of our knowledge, few of them utilize leaky-integrated discrete-time continuous-value neurons in the reservoir layer. Having leaky-integrated neurons is essential when dealing with temporal data as it controls the speed of the reservoir update dynamics and eventually the duration of the short-term memory of ESN~\cite{lukovsevivcius2012practical}. Besides the lack of using leaky-integrated neurons, none of the aforementioned implementations take into consideration the  limited precision of signal digitization and undesired leakage of the sample and hold circuits during network training and testing. This paper addresses these problems with key contributions summarized as follows:
\begin{itemize}
[noitemsep,topsep=1pt]
    \item Introducing an energy-efficient memristive ESN accelerator, targeting AI-powered edge devices.
    \item Enhancing the hardware-aware adaptation and the dynamic diversity of ESN via enabling in-situ learning (synaptogenesis and synaptic plasticity) and using leaky-integrated neurons.
    \item Investigating the potential of enhancing the lifespan of ESN and its robustness to device failure.
    \item Evaluating the proposed accelerator for robustness, latency, network lifespan, and energy-delay product.
\end{itemize}

The rest of the paper is organized as follows: Section II discusses the theory of ESN network and training procedure. Section III presents the system design and implementation of the ESN. The design methodology is introduced in Section IV. Sections V and VI discuss the results and conclude the paper, respectively.

%%%%%%%%%%%%%%%%%%%%%%%%%%%%%%%%%%%%%%%%%%%%%%%%%%%%%%%%%%%%%%%%%%%

\begin{figure}[!t]
\begin{center}
\includegraphics[width=0.55 \textwidth]{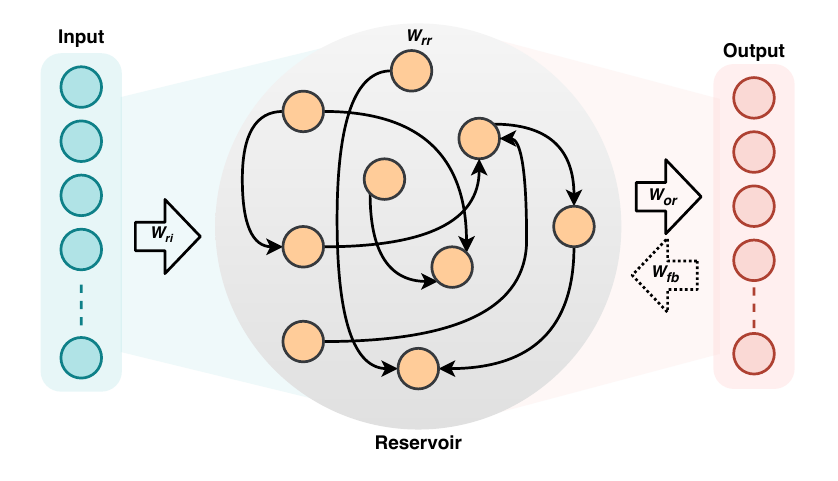}
\caption{A high-level diagram of the ESN network which consists of an input, reservoir, and readout layer. The input works as a buffer, whereas the reservoir and readout layers are dedicated to feature extraction and classification, respectively.}
\label{esn}
\end{center}
\end{figure}

\section{Overview of Echo state Network}
The ESN is a class of reservoir networks originally proposed by Jaeger~\cite{jaeger2001echo} in 2002 to circumvent challenges associated with training recurrent neural networks (RNNs) (vanishing and exploding gradient when using backpropagation through time~\cite{gauthier2021next}). The algorithm is inspired by the structure of the neocortex and attempts to model its dynamics and sequence learning~\cite{zhang2023survey}. The high-level diagram of the ESN, shown in~\fig{esn}, comprises three consecutive layers of $n_u$ input neurons, $n_r$ reservoir (or hidden) neurons, and $n_o$ readout (or output) neurons. The input layer serves as a buffer, holding the information presented to the network. The reservoir works as a feature extractor that provides non-linear dynamics and memory, enabling the network to process information of a temporal nature. The output layer is a linear classifier. When the ESN algorithm is used for time-series forecasting, it operates in three phases: initialization, prediction, and training, summarized in Algorithm-I. During the initialization phase (Lines 2-5), the strength of the synaptic connections (weights) is randomly initialized and the features associated with the structure of the network, such as the reservoir feedback connections, are set. Besides the synaptic initialization, all non-zero elements of the reservoir weight matrix are scaled in width to get the network operating at the edge of chaos. Once the initialization phase is finished, which occurs only once, the prediction phase begins. Here, the input examples ($U = \{\vec{u}^{1}, \vec{u}^{2}, ...., \vec{u}^{n_m}\}$) are presented to the network in an online fashion. Given an example, $\vec{u} \in \mathbb{R}^{n_u \times n_t}$, its features are sequentially presented to the network where they initially get multiplied by the synaptic weights connecting the reservoir and input layer, $W_{ri} \in \mathbb{R}^{n_r \times n_u}$, and then summed up. Concurrently, within the reservoir layer, the outputs of the reservoir neurons from the previous time step ($x^{<t-1>}$) are multiplied by their corresponding feedback connection weights, $W_{rr} \in \mathbb{R}^{n_r \times n_r}$, and summed up. The weighted sums from both input and feedback connections are added and then subjected to a hyperbolic tangent \textit{tanh} activation function ($g$) to introduce non-linearity. The output of the \textit{tanh} function, $\hat{x}^{<t>}$, represents the reservoir neuron's internal state, which is used along with the leakage rate ($\delta$) to determine the final output of the reservoir neurons, $x^{<t>}$, see [Lines 8-9]. The reservoir neurons in this work capture the leaky-integrated discrete-time continuous-value feature to ensure broader control over the reservoir update dynamics. The outputs of the reservoir neurons are then transmitted to the readout layer via the weighted synaptic connections, $W_{or} \in \mathbb{R}^{n_o \times n_r}$, and the weighted sum is subsequently fed to a sigmoid activation function ($f$) to compute the predicted value. Once the predicted value is computed, the training phase starts to minimize the error between the predicted and targeted values (for forecasting, the targeted value is $u^{<t+n_p>}$ in the time-series data, where $n_p$ denotes the number of prediction steps). Unlike other networks, the training in the ESN is confined solely to the output layer. This makes the algorithm well-known for its fast training, thereby making it attractive to numerous applications. Various training algorithms can be used to train the ESN, such as backpropagation-decorrelation, force learning~\cite{sussillo2009generating}, ridge regression, etc., but the most common approach is ridge regression~\cite{lukovsevivcius2012practical}. Ridge regression involves finding the optimal set of weights that minimize the squared error between the outputs of the network ($\hat{Y}$) and ground truth labels ($Y$). It is carried out by multiplying the Moore-Penrose generalized inverse of the reservoir output ($X$) by ground truth labels. Orthogonal projection and Tikhonov regularization ($\lambda$) are recommended to be employed here to overcome overfitting and enhance stability, see \eq{ridge}. 

\begin{equation}
    W_{or} = YX (XX^T + \lambda I)^{-1}
\label{ridge}
\end{equation}

\setlength{\textfloatsep}{3mm}
\IncMargin{1em}
\begin{algorithm}[!t]
\caption{ESN Algorithm}
\label{alg:one}
\KwIn{$\vec{u} \in \mathbb{R}^{n_u \times n_t}_{[-1,1]}$, where $\vec{u} \subset U$ and $U \in \mathbb{R}^{n_u \times n_m \times n_t}_{[-1,1]};$~~~~~~~~~~~~~~~~~~~~~~~~~~~~~~~~\scriptsize{\tcc*[h]{$n_m$: Number of examples}}
}
\KwOut{$\vec{\hat{y}} \in \mathbb{R}^{n_c \times (n_t - n_p)}_{[0,1]}$\scriptsize{\tcc*[r]{$n_c$:Predicted output dim.}}} 
// Initialization:~~~~~~~~~~~~~\\
$W_{ri} \sim$ rand.uniform, where $W_{ri} \in \mathbb{R}^{n_r\times n_{u}}_{[-1, 1]}$ ;~~~~~~~~~~~~~~~~~~~~~~~~~~~~~~~~~~~~~~~~~~~~~~~~{\scriptsize\tcc*[h]{$n_u$: Input features count}}\\
$W_{rr} \sim$ rand.uniform.sparse, where $W_{rr} \in \mathbb{R}^{n_r\times n_r}_{[-0.1, 0.1]}$ \;
$W_{or} \sim$ rand.uniform, where $W_{or} \in \mathbb{R}^{n_o\times n_r}_{[-1, 1]}$ \;
%$W_{rr} \leftarrow  \frac{W_{rr}}{\psi({W_{rr}})}$;~~~~\tcc*[h]{\scriptsize{$\psi$ is the max. abs. eignevalue}} \\	
$count \leftarrow 1$ \;
\Repeat{$t > n_t-n_p$}
{
	// Prediction phase\\
 	$\hat{x}^{<t>} \leftarrow g(W_{ri} \odot u^{<t>} + W_{rr} \odot x^{<t-1>})$; ~~~~~~~~~~~~~~~~~~~~~~~~~~~~~~~~~~~~~~~~{\scriptsize\tcc*[h]{$g$ is a tanh activation function}} \\			
 	${x}^{<t>} \leftarrow (1-\delta)~ x^{<t-1>} + \delta~ \hat{x}^{<t>}$ \;
 	$\hat{y}^{<t>} \leftarrow f(W_{or} \odot x^{<t>})$ ;~~~~~~~~~~~~~~~~~~~~~~~~~~~~~~~~~~~~~~~~~~~~~~~~~~~~~~~~~~~~~{\scriptsize\tcc*[h]{$f$ is a sigmoid activation function}} \\	
 	// Learning phase:\\
 	\If{Learning == 'Enabled'} 
 	{
  		$Er^{<t>} \leftarrow \hat{y}^{<t>} - y^{<t>}$\;
        $grad \leftarrow grad + x^{<t>} \otimes Er^{<t>}$; \\
        \If{count \% $n_{up} == 0$}
        {
            $grad[i] \leftarrow 0$~{when}~ $grad[i] < \Theta, \forall i$ ; \\
            $W_{or} \leftarrow W_{or} - \alpha~\frac{grad}{n_{up}} + \lambda~W_{or}$ \; 
            $grad \leftarrow 0$ ; 

        }
 	}
    $count \leftarrow count +1$ ;
}
\end{algorithm}
\DecMargin{1em}

While such a training approach is effective and recommended for training ESN, it is computationally costly and requires massive memory to store thousands of labels and states (reservoir outputs). One may resort to the recursive least squares (RLS) to mitigate the computational and memory requirements, but this approach is not feasible for edge devices with stringent resource constraints. It is important to highlight that the aforementioned training approaches suffer from a divergence problem when not trained for several time steps, generally caused by the accumulation of small errors. Such a problem is inevitable and cannot be avoided by altering the structure of the network, but can be mitigated by bringing the reservoir dynamic to its original state by representing ground truth input to the reservoir~\cite{moon2019temporal}. However, such an approach may work when dealing with stationary data but not real-world data of a non-stationary nature. Therefore, to overcome the challenges associated with finding the optimal set of parameters that ensure the best network performance while securing fast learning and adaptation, we use the online least mean square (LMS) algorithm due to its simplicity and minimal use for computational resources and storage. The LMS is used with weight decay regularization (LMS+L2) to optimize ESN performance even further as it is exposed to new events, see [lines 13-20]. Besides regularization, the gradients are sparsified [Line 16] to avoid insignificant changes in the network parameters and to enhance the learning process in hardware. Furthermore, the weights are tuned frequently according to the parameter $n_{up}$, rather than every iteration. This will expedite network adaptation and minimize energy consumption.

% ==========================================
\section{System Design and Implementation}
\begin{figure*}[!ht]
\begin{center}
\includegraphics[width=1 \textwidth]{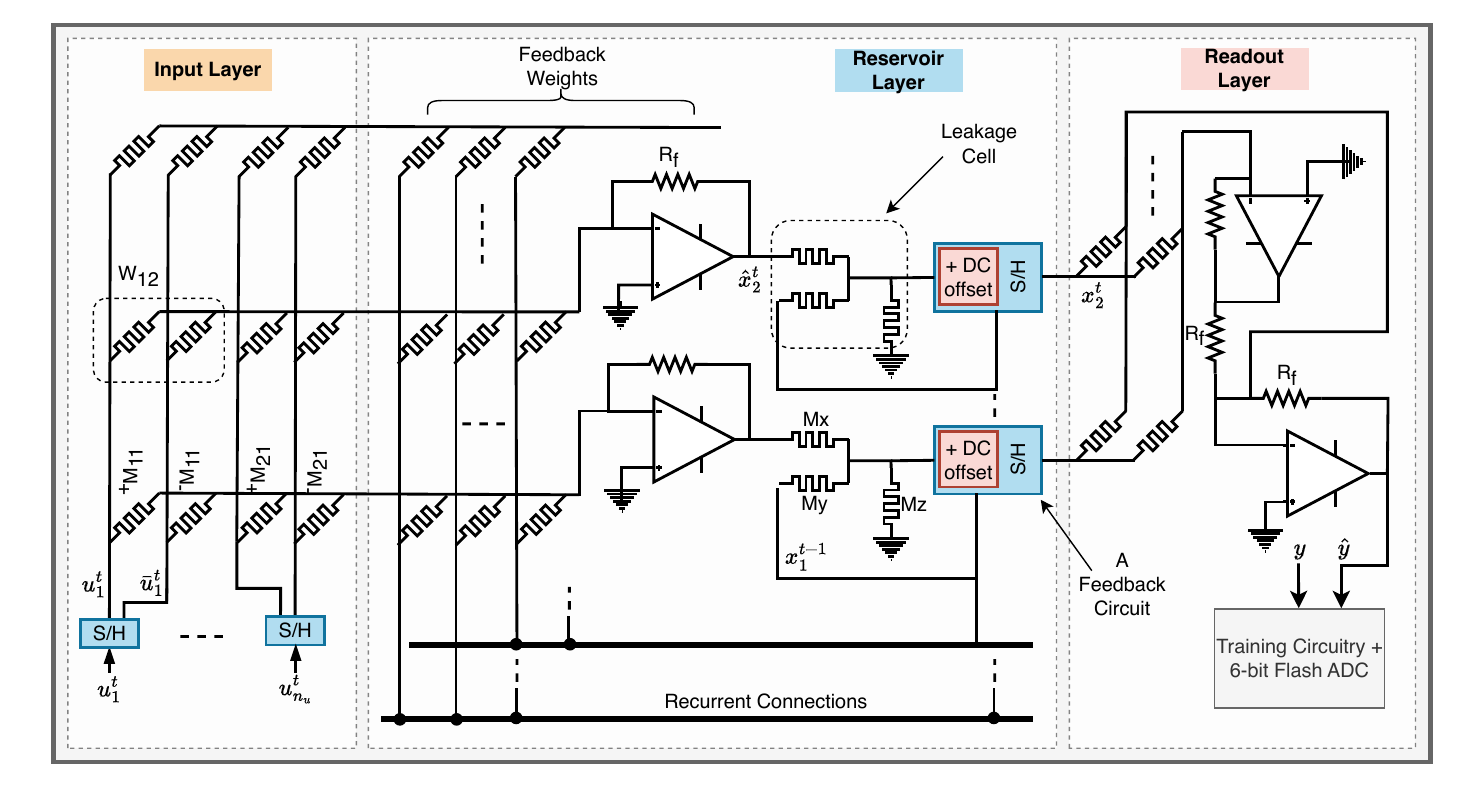}
\caption{The schematic of the developed ESN accelerator, including input, reservoir, and readout layers. Each layer comprises sample-and-hold circuits to discretize and temporally hold the time-series data, memristor devices to emulate the synaptic weights, and neuron circuits (leaky-integrated discrete-time continuous-value neurons in the reservoir and point neurons in the output layer). In the output layer, an additional unit, the training circuitry, is employed to enable in-situ learning.}
\label{reservoir_layer}
\end{center}
\end{figure*}

The general architecture of an ESN, as covered in the overview section, is composed of consecutive layers of fully and sparsely connected neurons. In hardware, the same architecture is captured, but inter-layer connections are buffered to control data flow (see~\fig{reservoir_layer}). During the forward pass, the input signals are sampled and held, and then presented to the reservoir layer via weighted connections modeled by memristor devices. Once the weighted sum of the input reaches the reservoir neurons, collectively with weighted reservoir neuron responses from the previous time step, the reservoir neuron non-linear response is determined. The output of the reservoir is classified using the readout layer, which is trained in an online fashion as alluded to earlier. 

The input layer is equipped with a sample-and-hold (S/H) circuit\footnote{To ensure high energy efficiency and a wide input range, we used the S/H circuit presented in~\cite{o200410}. The circuit is modified to generate a differential output and to hold multiple clones of the sampled input, which will be used during the prediction and training phases.} to discretize the continuous time-series input signal and to hold it until the reservoir neurons are ready to receive a stimulus. The stimulus to the reservoir neurons is presented via a set of memristor devices integrated into the crossbar structure. These memristors model the synaptic weights bridging the input-reservoir layers. Using one memristor to model each synaptic weight in the network is not common, as it can solely model unipolar weights. Thus, two cross-coupled memristors (2M structure) are typically used, where the net difference between their conductance results in a bipolar weight representation. 
%One may have two possible scenarios when it comes to number of used memristors and integration approach. The first is called 1M1R, in which ($n_i \times n_r+1$) crossbar is used, where each memristor in the crossbar corresponds to one weight representation and the addition columns is reserved to the reference memristor. Hence, the net difference between the conductance of any tunable memristor and the corresponding fixed memristor results in bipolar weight representation~\cite{zyarah2018semi}. The second involves utilizing two memristors to model each synaptic weights thus 2x($n_i \times n_r$) crossbar is used. While the first approach is knwon 
Using two memristors doubles the resources utilized in each layer, leading to uncompact and a less energy-efficient design. To ensure efficiency, one may resort to a 1M1R structure. In 1M1R, ($n_u \times n_r+1$) crossbar is used, where each memristor in the crossbar corresponds to one weight representation and the additional column is reserved for the reference resistor (or untunable memristor). Hence, the net difference between the conductance of any tunable memristor and the corresponding reference resistor results in bipolar weight representation~\cite{zyarah2018semi}. While the 1M1R structure uses fewer resources, it suffers from a limited dynamic range and may not be favorable when implementing streaming networks due to memristor endurance limitations. In this work, we adopted the 2M structure and attempted to leverage network and device inherent features to minimize resource usage while enhancing network robustness and performance. 

\begin{equation}
\begin{aligned}\label{current}
\hat{x_i}^{<t>} = g[(\frac{R_f}{M^-_{i1}} - \frac{R_f}{M^+_{i1}})~ u^{<t>}+ (\frac{R_f}{M^-_{i2}} - \frac{R_f}{M^+_{i2}}) ~x^{<t-1>}_{1}  + .... + (\frac{R_f}{M^-_{i(n_h+1)}} - \frac{R_f}{M^+_{i(n_h+1)}})~x^{<t-1>}_{n_h}]
\end{aligned}
\end{equation}

\begin{equation}
x_i^{<t>} \simeq  \underbrace{\frac{M_z || M_y}{(M_z || M_y) + M_x}}_{\delta}\hat{x}^{<t>}+  \underbrace{\frac{M_z||M_x}{(M_z || M_x)+M_y}}_{1-\delta}x^{<t-1>}
\end{equation}

When the stimulus reaches the reservoir layer, the internal states of all the reservoir neurons, $\hat{x}^{<t>}$, are updated. This process begins by computing the sum of inputs ($u^{<t>}$ and $x^{<t-1>}$) weighted by non-tunable memristor devices (random-fixed weights), as described in~\eq{current}. Then, the non-linearity, represented by $tanh$ function, is approximated by leveraging the linear operating region and limited bias of pMOS-input operational amplifiers (Op-Amps). After determining the internal states of the reservoir neuron, the leaky-integrated feature is captured to assess how much information from the previous activation should be carried over to the current hidden state. Implementing this feature in hardware may involve using at least one additional Op-Amp, leading to high power consumption and a large footprint area. To circumvent this issue, we introduce the so called leakage cell. The leakage cell combines the internal state and the previous activations using three memristors connected as shown in~\fig{reservoir_layer}. These memristors model the leakage terms, with the internal state being multiplied by the leakage rate ($\delta$), modeled by $\frac{M_z || M_y}{(M_z || M_y) + M_x}$, and the previous reservoir neuron output being multiplied by ($1 - \delta$), modeled by $\frac{M_z||M_x}{(M_z || M_x)+M_y}$\footnote{The resistance of the $M_z$ memristor should be set to be really large as compared to $M_x$ and $M_y$ so that the approximated $\delta$ and $1-\delta$ always sum up to one.}. This approach reduces the power consumption and area of the reservoir layer by more than 2$\times$ and 170$\times$, respectively, compared to using Op-Amps.
%While the Op-Amps provide a virtual ground to the crossbar bit-lines, eliminating other factors impacting weight representation with memristors, using Op-Amps to approximate the non-linearity results in high power consumption and area footprint. To circumvent this issue, we exploit the virtual ground provided by the leaky integrated neuron circuit and use a diode-based limiter to approximate the non-linearity~\fig{tanhx}. This reduces the power consumption and area of the reservoir layer by 2$\times$ and 170$\times$, respectively.
% Crossbar area = (16x60)+(16x110) 
% Activation cct (Op-Amp) = 2op x (42x52)
% Activation cct (diodes) = 2diode x (1.5x1.5)
% Area = [(16x60+16x110)+(128x2x42x52)]/[(16x60+16x110)+(128x2x1.5x1.5)]
%%%%%%%%%%%%%%%%%%%%%%%%%%%%%%%%%%%%%%
% Power consumption of reservoir network when using 2 Ops is 150.4mW
However, once the final outputs of the reservoir neurons are ready, they will be relayed to the feedback circuit. In the feedback circuit, the activation of each reservoir neuron is bypassed to the output layer to compute the final network output, $\hat{y}$, and  sampled and held to be used as feedback in the upcoming iterations. Due to the fact that the activation of the reservoir neuron is a result of \textit{tanh} function, it can be either positive or negative. Storing a negative value in a S/H circuit is challenging and not feasible. To this end, we superimposed the output of each reservoir neuron on a fixed DC offset to bring all outputs to positive ranges during the sampling phase. The DC offset is canceled once the stored values are represented to the reservoir. Once the output of the reservoir layer is ready, it is multiplied by the corresponding weights, and the final output of the ESN is computed. Then the training phase starts, which will be discussed in detail in the next subsection.

% \begin{figure}[t!]
% \begin{center}
% \includegraphics[width=0.55 \textwidth]{Figures/Input_Reservoir3.pdf}
% \caption{Memristor-based reservoir neuron, in which dual-memristor devices are used to emulate bi-polar weights. The non-linearity of the neuron is approximated by: a) Op-Amps, b) diodes.}
% \label{input_res}
% \end{center}
% \end{figure}

%we split the activation of each reservoir output into sign and magnitude, which can be stored into a S/H circuit and 1-bit register, respectively. The S/H circuit uses fully-differential op-amp with reversible outputs controlled by the sign register. The reversible outputs here refers to the fact that the output terminals can be swapped (positive becomes negative and vise versa). This feature is enabled via using transmission gates. 

% \begin{figure}[b!]
% \begin{center}
% \includegraphics[width=0.45 \textwidth]{tanh_activation.pdf}
% \caption{Approximation of the \textit{tanh} activation function achieved by: i) levering the linear operating region and limited bias of pMOS-input 3-stage Op-Amp implemented in standard 65nm technology node, ii) using a diode-based limiter.}
% \label{tanhx}
% \end{center}
% \end{figure}

It is important to highlight that each crossbar in the reservoir layer uses a 1M configuration. Although this configuration is well-known to possess a 10$\times$ smaller footprint as compared to 1T1M~\cite{yakopcic2014memristor}, it is harder to configure. Reconfigurability is an essential feature in ESN hardware implementation as it endows the network with an additional degree of freedom to continuously change its structure and sparsity level, ensuring network stability. In this work, reconfigurability (synaptogenesis) is enabled through the use of the Ziksa writing scheme, which we previously proposed in~\cite{zyarah2017ziksa}. Ziksa is used to suppress the effect of the synaptic connections that are not involved in computation, promoting network sparsity. This suppression occurs by setting the conductance of the memristor devices emulating the positive portion of the weights ($M^+$) to be equal to that representing the negative portion of the weights ($M^-$), such that the net current flowing through these synapses for a given input is zero. Our approach significantly reduces the resources required to enable sparsity, unlike the work in~\cite{kume2020tuning}, which suggests using switching transistors and 1-bit memory cells or flip-flops to control each MOSFET device in the crossbar. 

\begin{figure}[b!]
\begin{center}
\includegraphics[width=0.6 \textwidth]{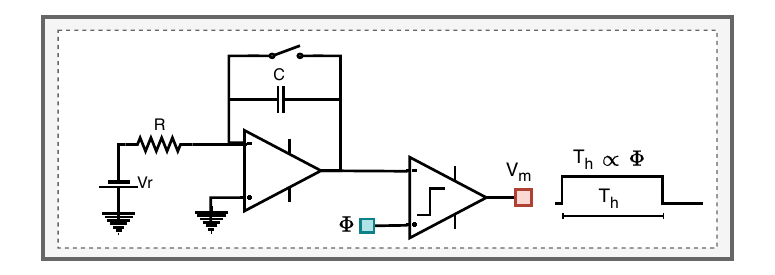}
\caption{The designed circuit used to convert ($\Phi$) into a pulse signal of fixed amplitude and variable duration. The circuit encompasses an integrator and comparator with internal positive feedback.}
\label{integrator}
\end{center}
\end{figure}

%Using pass transistors is preferred here to endow the network the necessary reconfigurability to change the reservoir sparse connections 
%Note: the pass transistors in the reservoir layer serve several advantages such as having reconfigurable random recurrent connections within the reservoir layer, suppressing the sneak paths. Additionally, in the output layer, the pass transistors are exploited by the training system to simplify the training system and to do more precise memristors tuning.  

\subsection{ESN Accelerator Training}
As discussed in the overview section, training in ESNs is confined solely to the readout layer, and it can be performed either analytically via ridge regression or iteratively by using LMS. In this work, we choose LMS to ensure fast learning and minimize resource usage. However, the LMS training equation consists of two terms: gradient and regularization. The gradient is first estimated by computing network output error, the difference between the predicted value and the ground truth as presented by the training circuitry block. Typically, a subtractor, a non-inverting operational amplifier, is used for this purpose. The output of the subtractor is multiplied by the activations of the reservoir neurons to estimate the gradient required to modulate the corresponding memristor conductance. The gradient is digitized using a 6-bit Flash-ADC (optimized for power constrained devices) and the output is sent to a global controller where it gets temporarily stored. Then, the regularization term is estimated as follows: First, a fixed test voltage\footnote{Due to technology limitations, the test signal ($V_{test}$) is set to be 50mV to avoid any undesired clipping or distortion in the generated output. } is applied to the memristor that needs to be updated ($M_u$). The output voltage is then captured again by the ADC, and the result will be sent to the global controller to estimate the value of the memristor conductance according to \eq{reg_mem}.

\begin{equation}
    M_u = R_f \times \frac{V_{test}}{V_o}
\label{reg_mem}
\end{equation}

Once the memristor conductance is estimated, it is multiplied by the regularization factor, and the result is added to the gradient in order to estimate the amount of change that needs to be made to a memristor. This amount ($\Phi$) is then translated into a pulse signal with a fixed amplitude and variable duration (duration reflects the amount of change to be made to a memristor) using the conversion circuit shown in~\fig{integrator}, where the output ($V_m$) is given by~\eq{integ2} and \eq{integ1}. Here, $V_{int}$ denotes the output of the integrator, and $V_r$ is a reference input voltage, which along with R and C, controls the speed of integration. 

\begin{equation}
V_m = 
\begin{cases}
T_h,& \Phi > V_{int} \\
0, & otherwise
\end{cases}
\label{integ2}
\end{equation}

\begin{equation}
T_h = \Phi \times \frac{V_r}{RC}
\label{integ}
\end{equation}

It is imperative to mention that the changes in memristors occur in a sequential fashion, one memristor at a time. This approach allows for more precise tuning and reduces power consumption during the training process. While such a training approach is known to be slow, we speed it up by means of gradient sparsification, which limits the changes to the devices with gradient values above a predefined threshold ($\Theta$). The training operation is verified in Cadence Virtuoso for a small-scale network (1x20x1)\footnote{The training is verified while considering memristor  cycle-to-cycle and device-to-device variabilities. Variabilities with a normal distribution has been considered, with a mean defined by device parameters and a standard deviation equal to 10\% of the mean.}, whereas the large-scale ESN verification is done using a Python model. The Python model takes into consideration all the hardware constraints, such as limited range and precision of the weight representation, non-linearity of the memristor devices, leakage in the sample and hold circuit, etc.

% =================================================
\section{Methodology}
\subsection{Design Space Exploration}
In order to select the best hyperparameters that result in a compact design and optimal performance, the particle swarm optimization algorithm (PSO) is used. The algorithm is integrated with the ESN software model and set into constrained mode, in which the feasible points for hyperparameters are defined based on the hardware constraints. This includes the limited range of weight representation, crossbar size, precision, etc. In this work, the PSO employs a swarm of 50 particles, which are chosen to optimize for the following: the number of reservoir neurons that controls the high-dimensional space of feature extraction, frequency of synaptic updates ($n_{up}$), sparsity level within the reservoir layer which influences the computational cost, leakage rate to moderate the speed of the reservoir dynamic updates, regularization term to mitigate the overfitting, and learning rate to regulate convergence speed\footnote{The leakage rate, regularization term, and learning rate are set to be global parameters.}. The algorithm runs for 30 iterations with the target of minimizing the wMAPE of real-world benchmarks when performing forecasting tasks.

\subsection{Device Non-idealities and Process Limitations}
There are several operational and reliability concerns when using memristor devices to model the synaptic connections in neural network accelerators, such as endurance limitations and device non-idealities. The endurance limitation is highly impacted by several factors, including the programming voltage, device structure, and device material. For instance, using excessive writing voltage can lead to a major deterioration in memristor endurance, whereas the proper use of programming voltage leads to a significant improvement~\cite{wang2015impact}. However, crossing the endurance limits may drive the memristor devices to exhibit various behaviors which may severely impact network performance. For example, the authors in~\cite{yang2010high} observed a reduction in the metal-oxide memristor (Pt/TaOx/Ta) resistance ratio once the switching cycles surpass $6\times10^9$. Once the device switching window is collapsed, the device will be stuck at the reset resistance rather than being shorted. In~\cite{kim2016voltage}, the authors observed different behavior. After surpassing the device switching cycles limit, the memristor experienced a stuck-at failure. In this work, since we are modeling the physical characteristics of the device proposed by~\cite{yang2010high}, we will consider that the memristors experience stuck-at failure when passing the endurance limit. The endurance limit considered is $1\times10^9$ cycles when tuned with a fixed voltage pulse of $\pm$1.2v for set and reset.

When it comes to the device non-idealities, we have cycle-to-cycle and device-to-device variabilities, which characterize the time-varying stability of memristors and their uniformity when integrated into a crossbar structure~\cite{vourkas2016memristor}. These non-idealities typically occur due to device material and imperfect manufacturing processes~\cite{amat2016memristive}. However, in this work, the cycle-to-cycle variability of the device resistance range has been considered and is modeled as a variation in the weight range. For device-to-device variation, it is applied to the memristor threshold and resistance (write variation). Here, the write variation refers to the variability in the rate of change in device resistance during the learning process, which is modeled by adding noise to the learning rule. Regarding the process variation and limitations, it is considered for the developed 6-bit flash ADC, comparators, and operational amplifiers\footnote{All analog units are tested for fabrication process, voltage supply, and ambient temperature variations using corner analysis.}.

\subsection{Device Model}
The memristor device model used in this work is VTEAM~\cite{kvatinsky2015vteam}, which is fitted to the physical device characteristics proposed in~\cite{yang2010high}. The model is given by \eq{mem_eq} and \eq{mem_eq2}\footnote{$k_off$, $k_{on}$, $\alpha_{on}$, and $\alpha_{off}$ are constants, and $v_{off}$ and $v_{on}$ are the memristor threshold voltages.}, where $G_{on}$ and $G_{off}$ represent the maximum and minimum conductance range of the memristor device. $w$ and $D$ denote the device state variable and its thickness, respectively. Here, the change in device conductance is non-linear and it is governed by the state variable change when a voltage surpassing the device threshold is applied across its terminals. To achieve better fitting, the Z-window function, proposed in~\cite{zyarah2019neuromemrisitive}, is used. The Z-window has a broad range of parameters to control device characteristics. Furthermore, it possesses attractive features, such as overcoming the boundary lock problem, scaling, non-symmetrical behavior, etc. The Z-window is given in~\eq{mem_eq3}, where $\delta$, $k$, and $p$ refer to the sliding level (over the x-axis), scalability factor, and falling curve slope as it approaches the boundaries of the device. To comply with the device characteristics, the technology node, and targeted application constraints, the following has been considered: i) the memristor offers a high conductance range, ii) the set and reset voltages are no more than 1.2v, iii) the device variability is limited to 10\%.

\begin{equation}
G_{mem} = \frac{w}{D} \times G_{on} + (1 - \frac{w}{D}) \times G_{off}
\label{mem_eq}
\end{equation}

\begin{equation}
\frac{\Delta w}{\Delta t} = 
\begin{cases}
k_{off}.\Big(\frac{v(t)}{v_{off}} - 1\Big)^{\alpha_{off}}.f_{z}(w),&0 < v_{off} < v \\
0, &v_{on} < v< v_{off} \\
k_{on}.\Big(\frac{v(t)}{v_{on}} - 1\Big)^{\alpha_{on}}.f_{z}(w),&v <v_{on} < 0
\end{cases}
\label{mem_eq2}
\end{equation}

\begin{equation}
\label{mem_eq3}
    f_z(w) = \frac{k[1-2 (\frac{w}{D} - \delta)]^p}{e^{\tau (\frac{w}{D} - \delta)^p}}
\end{equation}

\begin{figure}[ht!]
\begin{center}
\includegraphics[width=0.45 \textwidth]{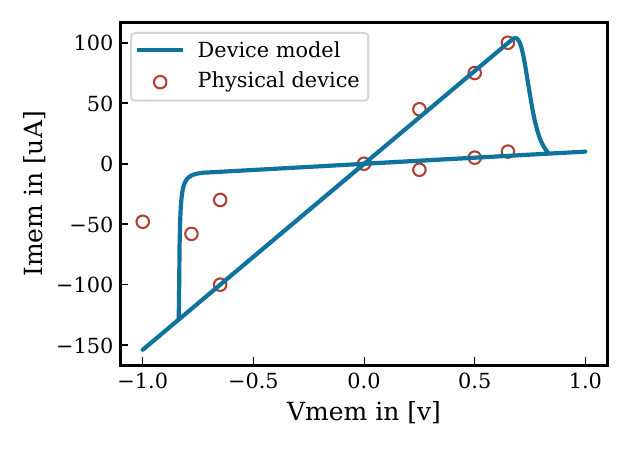}
\caption{Fitting of the used memristor device model to the physical device characteristics provided in~\cite{yang2010high}}.
\label{mem_char}
\end{center}
\end{figure}

\begin{table}[th!]
\small
\caption{The memristor device parameters used in the developed ESN accelerator.}
\label{mem_para_table}
\begin{center}
\begin{tabular}{lcc}
  \hline                       
\rowcolor{Gray}
\textbf{Parameter} & \textbf{Value [Reservoir and Readout]} & \textbf{Value [Leakage Cell]} \\ \hline
Memristor range & 200k$\Omega$ - 2M$\Omega$  & 100k$\Omega$ - 10M$\Omega$\\ 
Memristor threshold      & $\pm$1v & $\pm$1v \\ 
Full switching pulses    & 41 & 67 \\ 
Training voltage        & $\pm$1.2v & $\pm$1.2v\\ 
Endurance        & $1\times 10^{9}$ & $1\times 10^{9}$ \\
Switching time & $<$10 ns & $<$10 ns \\
\bottomrule
\end{tabular}
\end{center}
\end{table}

\subsection{Verification Benchmark}
In order to verify the operation of the proposed memristor-based ESN accelerator, several univariate benchmarks have been used in the time-series forecasting task:
\begin{itemize}
    \item PJM Energy: This dataset holds 145,366 samples representing regional energy consumption in the USA~\cite{PJM_Energy_Dataset}. The energy consumption is recorded every hour for the period 01/01/2010 to 07/01/2018. The ESN is employed for short-term energy consumption forecasting, i.e., the energy consumption for the next 50-100 hours. 
    
    \item Daily temperature: The daily minimum temperature in Melbourne, Australia recorded between the period (1981-1990)~\cite{daily_temp}. The dataset contains 3,605 noisy samples. We smoothed the data with moving average (window size=5) to reduce the noise effect and used ESN to predict the temperature for the next day.
    
    \item Mackey-Glass: This dataset is generated by non-linear time-delayed differential equation modeling a chaotic system~\cite{mackey1977oscillation}, see~\eq{mackey_eq}. It is widely used as a benchmark for time-series forecasting tasks. In this work, the parameters of the Mackey-glass equation are set as follows: $\beta = 0.25, \gamma=0.1, \tau = 18,~\text{and}~ n=10$, whereas the number of the generated examples used in forecasting task with ESN is 4000. 
    \begin{equation}
        \frac{dx}{dt} = \beta \frac{x(t-\tau)}{1+[x(t-\tau)]^n} - \gamma x(t)
    \label{mackey_eq}
    \end{equation}

    \item NARMA10: The classical non-linear autoregressive moving average (NARMA) was introduced by~\cite{atiya2000new}. It represents a dynamical system that is difficult to model due to its non-linearity and long-term dependencies. In this work, 4000 samples from the $10^{th}$ order NARMA system are employed for the forecasting task. The system is defined by~\eq{narma}, where $y(t)$ and $s(t)$, respectively, are the output and the input of the system at time $t$.
    \begin{equation}
    \begin{aligned}
     &y(t+1) = 0.3y(t) + 0.05y(t)\sum\limits^{9}_{t=0} y(t-i) + 1.5s(t-9)s(t)+0.1
    \end{aligned}
    \label{narma}
    \end{equation}
    
\end{itemize} 
All the benchmark features are scaled to range between 0 and 1 so that they have the same interval as the reservoir and output neurons. Furthermore, the first 100 samples are allocated to the initial washout period.

% =========================================

\section{Experimental Results and Discussion}
\subsection{Time-series Forecasting}
In order to quantify the performance of the proposed ESN memristor-based accelerator, time-series forecasting is evaluated using the weighted mean absolute percentage error (wMAPE) metric, which is given by~\eq{mape}. With weights ($\varpi _i$) set to unity, the wMAPE can determine the average difference between the actual and predicted values while capturing the substantial fluctuations in magnitude~\cite{miraftabzadeh2023day}.

\begin{equation}
\centering
wMAPE = \frac{\sum\limits^{0.5 \times n_t}_{t=1} \varpi_i|y^t - \hat{y}^t|}{\sum\limits^{0.5 \times n_t}_{t=1} \varpi_i|y^t|}
\label{mape}
\end{equation}

\fig{accur_th}-(a) depicts the wMAPE of the software and the developed hardware model of the ESN when using the PJM energy dataset to predict the energy consumption of the load for the next 50 hours. It is evident that the wMAPE, which is recorded every time 250 samples are introduced to the network, starts with a high value and then gradually degrades over time as the network learns and captures short and long-term dependencies. In the absence of the regularization term in the learning equation, it appears that the network takes a longer time to adapt. Furthermore, the network struggles to re-adapt when there is a major change in input patterns. When it comes to the developed hardware model, which uses regularization, a similar trend as the software model has been observed. However, one may notice there is a gap in performance ($\sim$3.9\%) between the software and hardware model, which can attributed to several reasons. Among these are the non-idealities of memristor devices, limited precision of the ADC used when modulating the conductance of the memristors, and undesired leakage in the sample and hold circuit leading to a drop in the stored charges.

\begin{table}[t!]
\small
\caption{The forecasting wMAPE of univariate stationary and non-stationary benchmarks using the proposed memristor-based ESN when trained with LMS, LMS with L2 regularization, and when using point and leaky-integrated neurons (LINs) in the reservoir layer. }
\begin{center}
\begin{tabular}{*5c}
  \hline                       
\rowcolor{Gray}
 \textbf{Benchmark} & \textbf{Forecast}  &  \textbf{LMS (LIN)} &\textbf{LMS+L2}  & \textbf{LMS+L2 (LIN)}  \\
\midrule

PJM-Energy 
& 50-Step   & {0.073$\pm$0.0011} & {0.087$\pm$0.0024}   & {0.061$\pm$0.0084}\\
& 100-Step   & {0.075$\pm$0.010} & {0.092$\pm$0.023}   & {0.066 $\pm$0.0089}\\
\midrule
Mackey-Glass & 50-Step   & {0.053$\pm$0.0069} & {0.079$\pm$0.0273}  & {0.047$\pm$0.0004}  \\
& 100-step   & {0.060$\pm$0.011} & {0.082$\pm$0.033}   & {0.047$\pm$0.0067}\\
\midrule
Daily-Temp & 50-Step   & {0.075$\pm$0.0052} & {0.097$\pm$0.0189}  & {0.073$\pm$0.0015}\\
& 100-step   & {0.093$\pm$0.029} & {0.0105$\pm$0.025}   & {0.083$\pm$0.0172}\\
\midrule
NARMA10 & 50-Step   & {0.191$\pm$0.0039} & {0.198$\pm$0.0069}  & {0.189$\pm$0.0034}\\
& 100-Step   & {0.195$\pm$0.0039} & {0.211$\pm$0.0192}  & {0.191$\pm$0.0039}\\
\bottomrule
\end{tabular}
\end{center}
\label{mape_bench}
\end{table}

\tb{mape_bench} shows the wMAPE when forecasting data from stationary and non-stationary benchmarks for various time steps. The wMAPE is calculated when training the proposed memristor-based ESN accelerator with LMS, and LMS with L2 regularization using point and leaky-integrated neurons in the reservoir layer. It can be seen that having leaky-integrated neurons (LINs) significantly improves network performance as it enables the ESN to form temporal dependencies with controllable dynamic updates.

% Error = 100 - (100x0.093) 
% Maximum error is 100%
% Ratio = 9.3% - 8%
\subsection{Device Failure Effect}
\begin{figure*}[h!t]
\centering
\subfigure{\includegraphics[width=45mm, height=36mm]{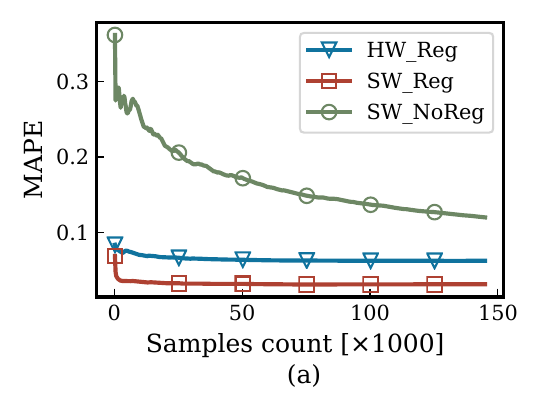}}
\subfigure{\includegraphics[width=45mm, height=36mm]{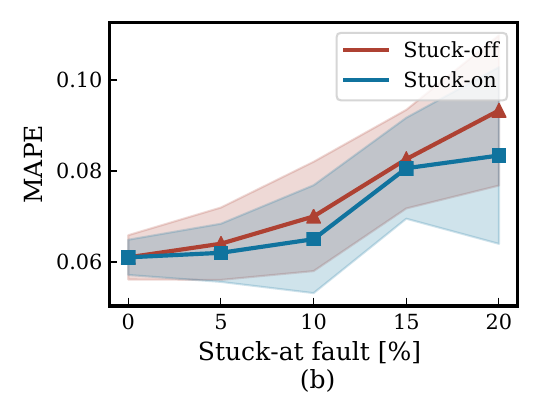}}
\subfigure{\includegraphics[width=45mm, height=36mm]{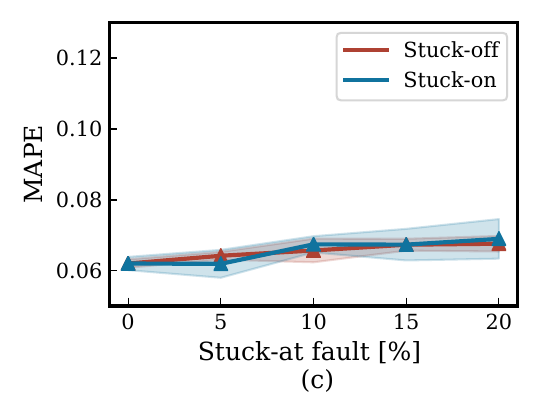}}
\caption{(a) The wMAPE of the developed ESN models (SW and HW) is calculated while predicting the load energy consumption for 50 hours, computed every 250 samples. (b) and (c) show the impact of various levels of stuck-on and stuck-off faults in memristor devices on the network performance while performing time-series forecasting task when using 1M1R and 2M crossbar structures, respectively.}
\label{accur_th}
\end{figure*}

There are several types of device failure a network has to deal with especially when using memristor devices. The memristor devices may experience a deviation in characteristics such as a change in the device conductance range due to excessive device switching, known as an aging fault, or it may have a fault in the device right after the fabrication process such as stuck-at-fault. In this work, we will primarily focus on stuck-at fault because it is most common and may severely impact network performance~\cite{xu2018optimized}. There are three types of stuck-at fault: stuck-at, stuck-on, and stuck-off. Previous work has shown that stuck-at replicates the memristor device failure after the forming process and such fault has marginal effect on network performance~\cite{zyarah2019neuromemristive}. Thus, this work solely investigates the impact of stuck-on (high-conductance state) and stuck-off (low-conductance state) faults on predicting future events in time-series data. In~\fig{accur_th}, we show the change in wMAPE when using the PJM energy dataset to predict the energy consumption of the load for the next 50 hours in the presence of various levels of stuck-on and stuck-off faults occurring in the readout and reservoir layers. The fault impact is investigated in both 1M1R and 2M crossbar structure. In 1M1R structure, it is found that regardless of the fault type, below 10\% its impact can be deemed marginal causing approximately $\pm$1.3\% change in the wMAPE value. However, this impact exacerbates beyond 10\% and seems to have an exponential negative impact on wMAPE, see~\fig{accur_th}-(b). In 2M structure, we may observe similar impact, but it can be suppressed by leveraging the programmable nature of memristor devices. When using two memristors to represent each synaptic weight, one can: i) leverage the intact memristors to tune the weights modeled by faulty devices and here the weights will not be frozen but rather have limited dynamic range, ii) enforce sparsity, i.e. zero weight values, via equating the conductance of the intact and faulty devices.~\fig{accur_th}-(c) illustrates the wMAPE in the presence of stuck-on and stuck-at faults when we enforce sparsity.

%%%%%%%%%%%
\subsection{Network Lifespan}
The network lifespan (LSP) is defined by its ability to sustain learning and acquire new knowledge~\cite{zyarah2020neuromorphic}. In memristor-based architectures, the lifespan is highly affected by the memristor's limited endurance, leading to either a gradual or severe drop in the dynamic range of network parameters, loss of elasticity, and eventually network performance. Typically, memristor devices, particularly oxide-based devices, offer an endurance ranging between $10^{6} - 10^{12}$~\cite{coll2019towards}. This low endurance is sufficient to get the network to continuously learn and update, but not for a long period of time. 

Estimating the lifespan of the memristor-based network is not a trivial task as it is influenced by several factors, such as memristor endurance variability, changing input statistics, network convergence time, etc. However, in this work, first-order estimation for the lifespan will be considered as given in~\eq{lifespan}, where $E_d$ and $\sigma$, respectively, are the device endurance and variability. $U_f$ denotes the update frequency of the memristor devices while learning:

\begin{equation}
    LSP = \frac{E_d \pm \sigma}{U_f}
\label{lifespan}
\end{equation}

For the used benchmarks, the PJM energy dataset, the input samples are recorded and presented to the network every hour. Thus, the lifespan of the proposed ESN accelerator, when trained on the PJM energy dataset, is $\sim$115,740 years, given $E_d=10^{9}$. However, this number can drop linearly with any increase in the update frequency. For instance, in Mackey-Glass if the update occurs every 100ms, the lifespan of the same network drops to 3.21 years. Thus, we suggest the following techniques to enhance the lifespan: i) Use two memristors in differential configuration to model the synaptic weights. This not only results in a wide weight dynamic range, but it also helps extending the lifetime of the network, assuming the training process is conducted in an alternating manner. ii) Utilize an R-M configuration to limit the voltage drop across memristor devices during the forming process and conventional switching. Reducing the voltage drop can either be achieved by explicitly integrating the memristor with a proper resistor~\cite{kim2016voltage} or inherently in a 1T1M configuration. Integrating memristor devices into a crossbar structure is also expected to extend the device endurance due to the wire parasitic resistance~\cite{salahuddin2018era}. Due to the simplicity and effectiveness of the first approach, alternating tuning, it is adopted in the developed ESN accelerator. Consequently, $\sim$2$\times$ enhancement in the lifespan of the network is achieved.

% L = 1.218 + 6.913 + 5.044 + 1.218 + 4.85 + 26.59
\subsection{Latency}
\begin{figure*}[!t]
\centering
\subfigure{\includegraphics[width=45mm, height=36mm]{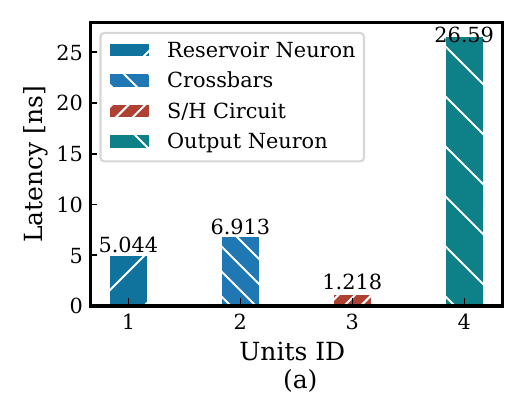}}
\subfigure{\includegraphics[width=45mm, height=36mm]{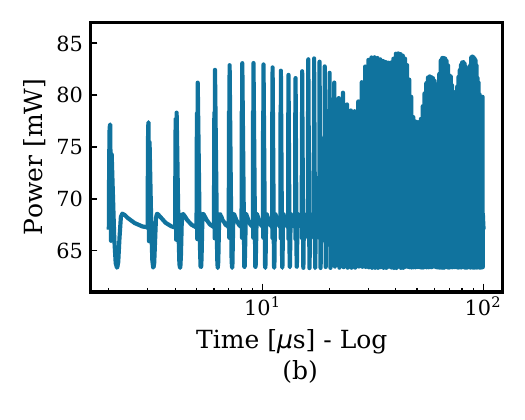}}
\subfigure{\includegraphics[width=45mm, height=36mm]{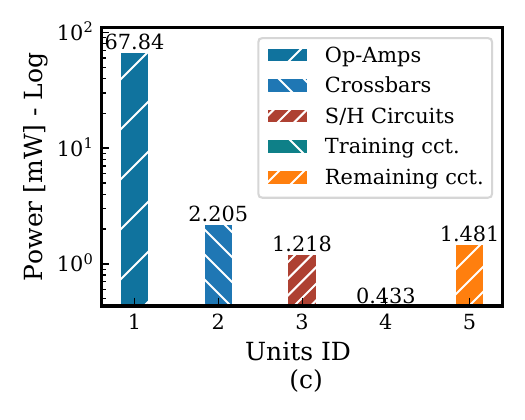}}
\caption{(a) and (c) The propagation delay and power consumption of the main circuit blocks used by the proposed memristor-based ESN accelerator. (b) The power consumption of the proposed accelerator recorded over time while forecasting future events from the PJM energy dataset.}
\label{power_laten}
\end{figure*}

The latency, the time required to process each input sample presented to the memristor-based ESN accelerator, is calculated while processing the univariate time-series dataset. During forecasting, the latency (worst-case scenario) is estimated to be 45.83ns, unequally contributed by several units in the network, see~\fig{power_laten}-(a). However, the units that account for$\sim$58\% of the delay are the neuron circuits, which are implemented using two Op-Amps to capture the non-linearity. This issue manifests only in the output layer, as each reservoir neuron uses one Op-Amp and a leakage cell to capture the non-linearity and leaky-integrated feature, respectively.

When comparing the presented memristor-based ESN accelerator with the digital counterpart clocked at 50MHz, $\sim$607$\times$ reduction in latency is witnessed when performing vector-matrix multiplication, owing to the extensive parallelism and in-memory computing of crossbar architecture, and $\sim$4$\times$ reduction in applying the non-linearity. It is important to mention here that to ensure precision when estimating the latency of the developed ESN accelerator, we built a large-scale network (1$\times$105$\times$1) in Cadence Virtuoso and estimated the time required to propagate an input signal through the individual components (developed under the 65nm process) and layers. The estimation of the propagation time is done using Cadence Virtuoso-ADE while considering resistance and capacitance parasitics, extracted in Mentor Graphics-Calibre.

% Training per column & 20us \\
% Tanh and LI Neuron & 5.044ns  \\
% Sigmoid & 26.59ns  \\

% Average power consumption of the entire network = 73.31mW
% Energy = 73.17mW x 45.83ns = 3.353nJ
\subsection{Energy-delay product}
The energy-delay product (EDP) of the proposed memristor-based accelerator implemented using a 65nm process is 153.68nJ$\times$ns. It is estimated in Cadence Virtuoso while forecasting the load energy consumption from the PJM energy dataset. It is important to note here that estimating the energy-delay product of a large-scale mixed-signal system tends to be challenging due to the disparity in the time-scale between simulation and actual hardware models. For instance, the samples of the PJM dataset are recorded every hour and estimating the energy consumption using the same time-scale can take days or weeks. Thus, to speed up the process, the time was scaled down by $2.7\times10^{-9}$.~\fig{power_laten}-(b) depicts the power consumption of the proposed ESN accelerator recorded over time. It can be observed that the static power is dominating as compared to the dynamic power consumption which manifests when input samples are presented to the network. In~\fig{power_laten}-(c), we report the breakdown of the average power consumption among the main units. One may observe that most of the power budget is directed towards the Op-Amps. Thus, in this work, we strive to reduce the number of the Op-Amp used in the reservoir layer, which results in more than 2$\times$ reduction in power consumption. 

It is essential to highlight that the joint enhancements in power consumption, latency, and architecture are reflected on the entire system leading to 247$\times$ reduction in energy consumption of the proposed memristive ESN accelerator compared to the digital counterpart~\cite{zyarah_esn_2023}. This makes the proposed accelerator more suitable for edge devices with stringent resources. \tb{HardwareAnalysis} provides a high-level comparison of the proposed ESN accelerator with the previous works. Our accelerator offers in-situ training, allowing for: i) processing of stationary and non-stationary data locally on the device, ii) hardware-aware learning and faster adaptation. Additionally, it features low latency as well as reasonable power consumption\footnote{No comparison in terms of energy or energy-delay product is conducted due to the lack of information in the references above.} as compared to other implementations reported in the literature.

\begin{table*}[!t]
\small
\caption{A comparison of the proposed ESN accelerator with previous work. One may note that these implementations are on different substrates, thereby this table offers a high-level reference template for ESN hardware rather than an absolute comparison.}
\label{HardwareAnalysis}
\setlength\tabcolsep{2.5 pt}
\begin{center}
\begin{threeparttable}
\begin{tabular}{lccccc}
\hline                      
\rowcolor{Gray} 
\textbf{Algorithm} & \textbf{Mixed-ESN \cite{kudithipudi2016design}} & \textbf{LS-ESN \cite{wen2018memristor}} & \textbf{Cyclic-ESN~\cite{liang2022rotating}} & \textbf{ESSM-ESN}~\cite{nair2023essm} & \textbf{This work} \\ \hline 
  Task   & Prediction    & Forecasting & Classification   & Classification & Forecasting  \\ 
  Reservoir size  & 30 & 100 & 8 & 128$\times$64$\times$28 & 105\\ 
  Input$\times$Output size  & 1$\times$1 & 1$\times$1 & 8$\times$5 & 76$\times$1 & 1$\times$1\\ 
  Power dissipation & 0.202mW  & - & 0.327mW\tnotex{tnote:robots-a3} & 58.38mW\tnotex{tnote:robots-a1} & 73.17mW\\ 
  Benchmarks &	ESD \& PF &	Load Power & Vowel Recognition & ECG  & PJM Energy \\
  Latency &  -      &     7.62s         &    50ns$<$  & -\tnotex{tnote:robots-a2} & 45.83ns   \\ 
Training     &  Off-Chip      &     On-Chip         &  Off-chip       & Off-Chip & On-Chip   \\
  Technology node     &  PTM 45nm      &     -         &  Standard 65nm       & PTM 22nm  & Standard 65nm   \\ \hline
  \end{tabular}
      \begin{tablenotes}
      \item\label{tnote:robots-a1} In~\cite{nair2023essm}, the power consumption when training the ESSM-ESN model is 2.04W. 
      \item\label{tnote:robots-a2} The latency is reported solely for the individual circuit blocks used in ESSM-ESN.
      \item\label{tnote:robots-a3} For the same network size and based on the provided information, the power consumption in~\cite{liang2022rotating} can reach $\sim$163.04mW when the latency is 50ns.  
    \end{tablenotes}
\end{threeparttable}
\end{center}
\end{table*}

%%%%%%%%
% =========================================
\section{Conclusions}
In this paper, we propose an energy-efficient memristor-based ESN accelerator to enable time-series data processing and learning on edge devices with stringent resources. The proposed design features in-situ training, enabling processing of stationary and non-stationary data, hardware-aware learning, and faster adaptation. When evaluated for time-series forecasting using standard benchmarks, it was found that the hardware model experiences a marginal degradation in performance compared to the software counterpart. This is attributed to several reasons, among which are the memristor devices' non-idealities, limited precision of the used ADC, and undesired leakage in the sample and hold circuits. Regarding the lifespan of the network, we suggest an alternating training approach to enhance the lifespan of the ESN. In the case of latency, while forecasting, we observe a significant reduction in latency compared to the digital implementations. This is because the most extensive operations, multiply-accumulate, are performed concurrently and in memory. When it comes to power and energy consumption, we notice that most of the power budget is directed toward the Op-Amps used to model the activation functions and capture the leaky-integrated feature. Thus, we strive to reduce the number of Op-Amps utilized to emulate the activation functions and capture the leaky-integrated features via introducing the memristor-based leakage cell. This results in more than 2$\times$ reduction in power consumption. The combined enhancements in architecture, latency, and power consumption give rise to a 247$\times$ reduction in energy consumption of the proposed memristive ESN accelerator compared to the digital counterpart implemented at the same technology node.

% \begin{acks}
% \end{acks}

\bibliographystyle{ACM-Reference-Format}
\bibliography{References}
\end{document}